# Reconfigurable Structural Robotic Assembly

## Interlocking 3D Aggregations with Self-Aligning Compound Nested Lattice Modules

**Alexander Htet Kyaw**
Massachusetts Instittue of Technology (MIT)

**Miana Smith**
Massachusetts Instittue of Technology (MIT)

**Paul Richard**
École Polytechnique Fédérale de Lausanne (EPFL)

**Neil Gershenfeld**
Massachusetts Instittue of Technology (MIT)

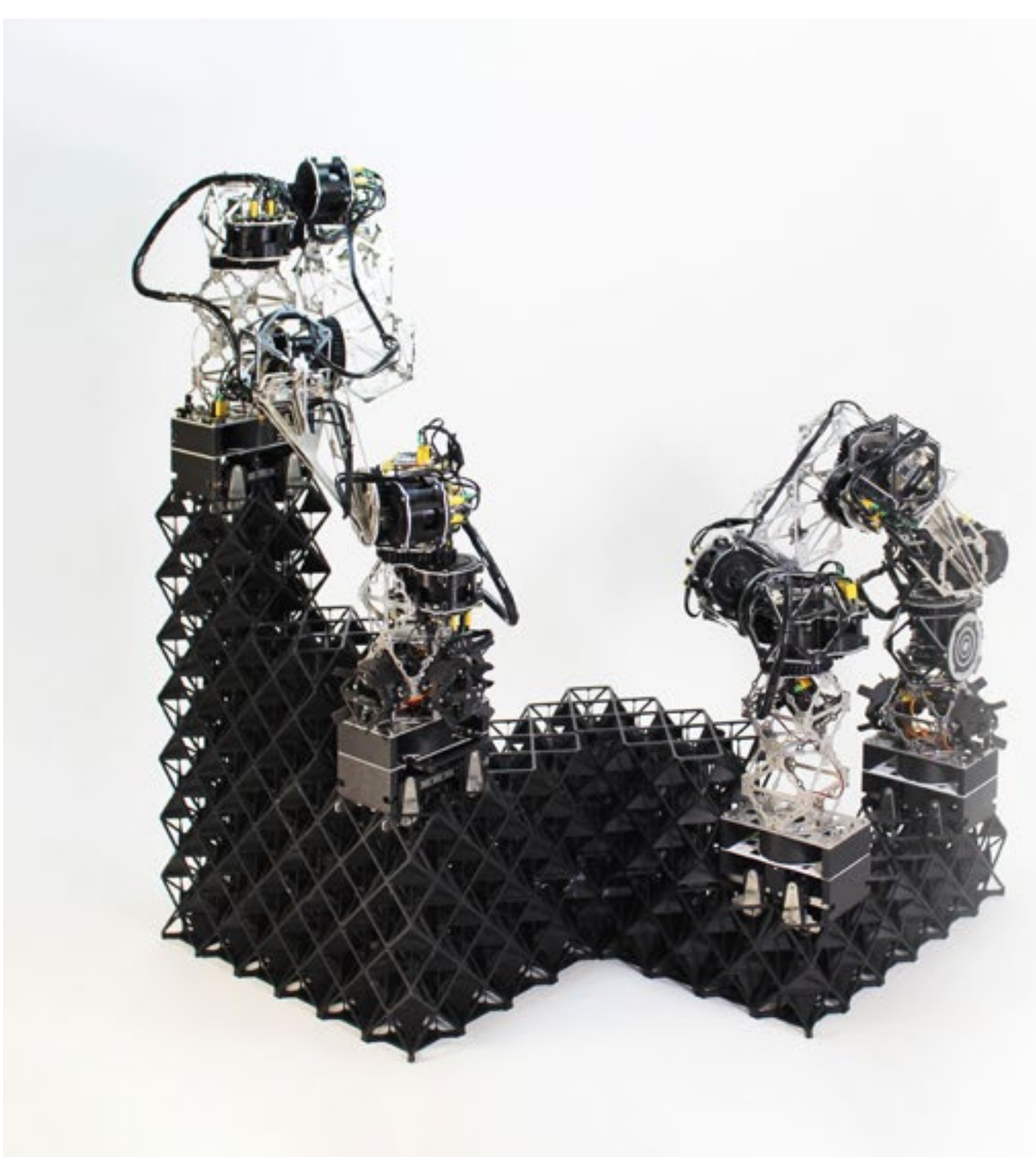

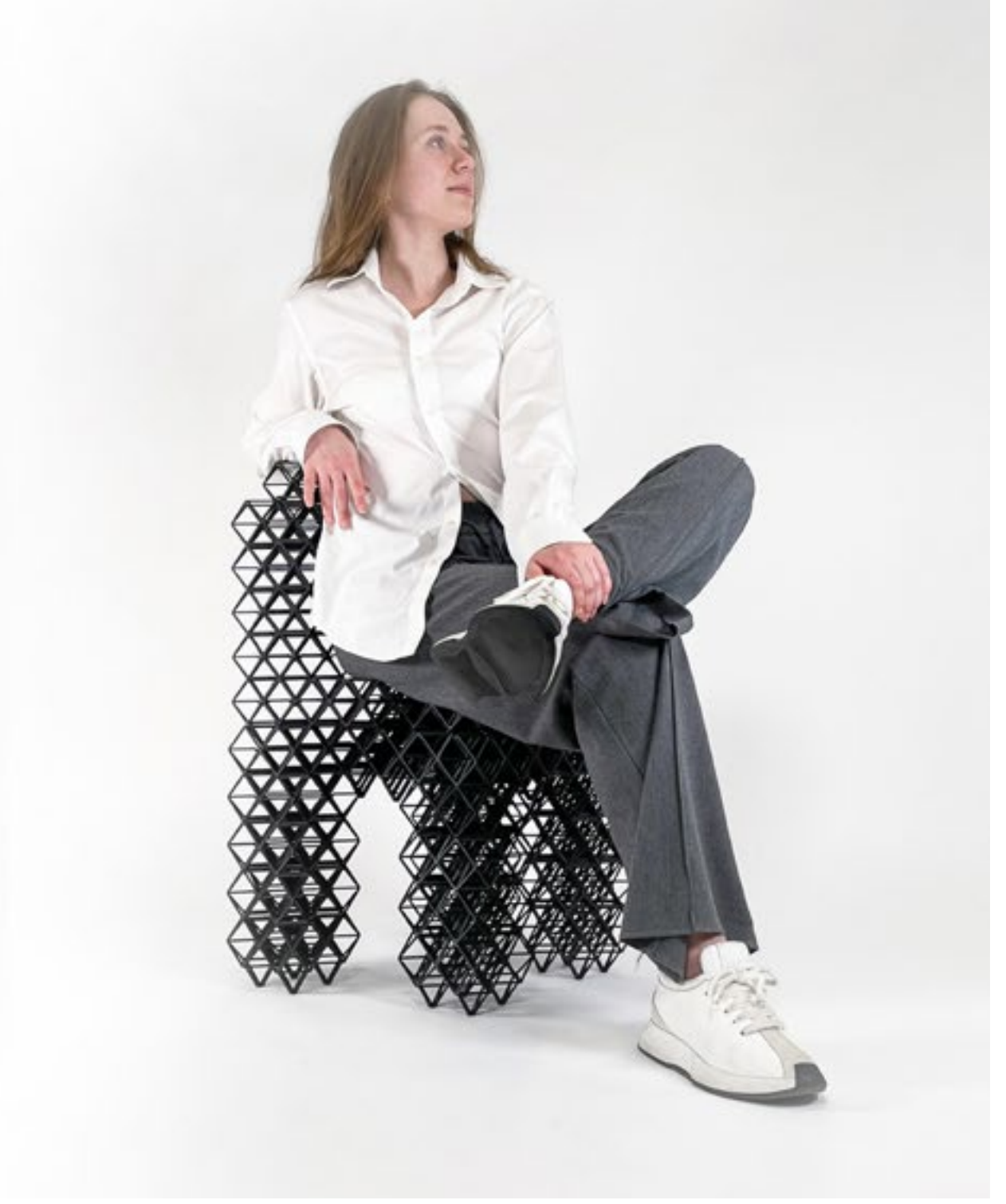

1

Robotic construction systems often treat the material system and the robot as separate design problems, locating intelligence primarily in hardware, sensing, motion planning, and control. This project instead investigates how geometric intelligence can be encoded within architected material systems to simultaneously address requirements for robotic grasping, self-alignment, reversible connection, structural performance, and three-dimensional aggregation.

We introduce a self-aligning compound nested lattice module composed of conjoined cuboctahedral–octahedral units. The cuboctahedral features of the modules provide defined surfaces for robotic grasping and alignment, while the octahedral features incorporate screw-releasable snap-fit connectors and corresponding receptors. Additionally, we present a nested arrangement that enables interlocking aggregation along the x, y, and z axes. We demonstrate the system through furniture and architectural scale structures assembled using both a robotic arm and mobile assembler.

The resulting configurations include seating, spanning structures, surfaces, and vertical frames. Compression testing of the compound module produced a stiffness of 4,556 N/mm, a maximum load of 3,445 N, and a compressive modulus of 17.5 MPa. The modules can also be disassembled and reused across different configurations, supporting reconfigurable and circular construction.

**PRODUCTION NOTES**

| | |
|---|---|
| Status: | Built |
| Site Area: | Various |
| Location: | United States |
| Date: | 2026 |

1 Mobile robot assembling a staircase structure using compound nested lattice units. By enabling distributed robotic assembly in parallel, mobile assemblers can construct larger structures beyond the reach of a single robotic arm or gantry-based system. An assembled load-bearing chair is shown on the right as another reconfiguration.

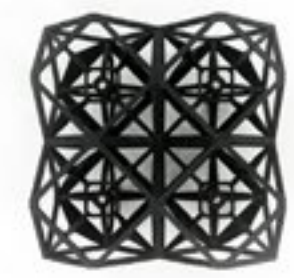
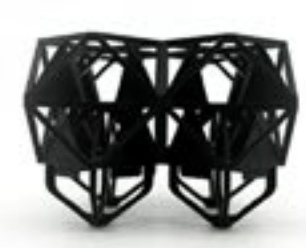

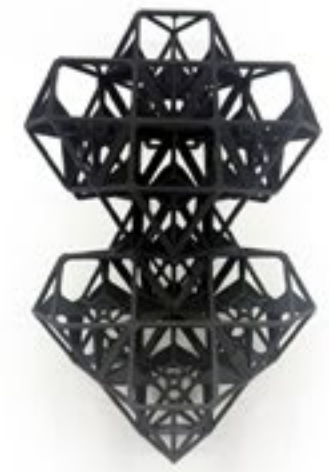
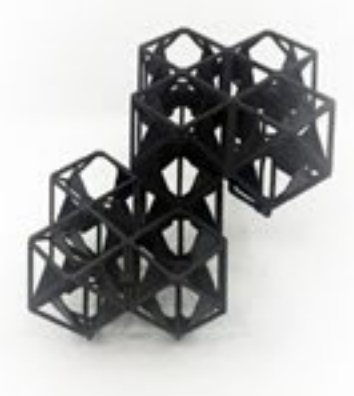

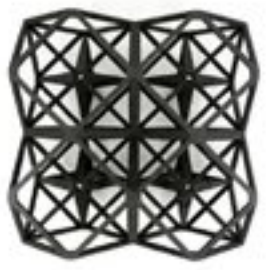
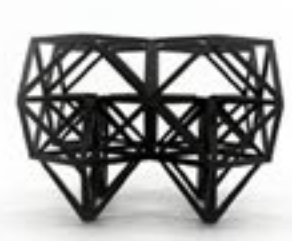

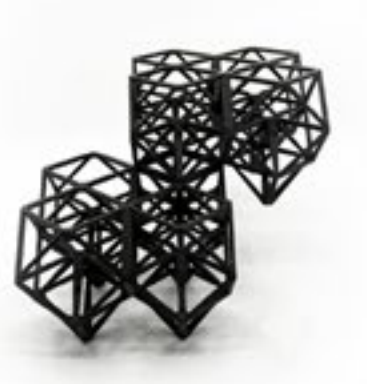
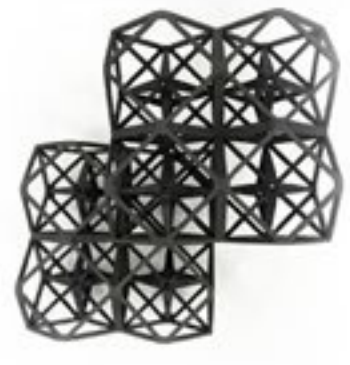

2

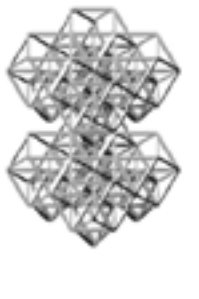
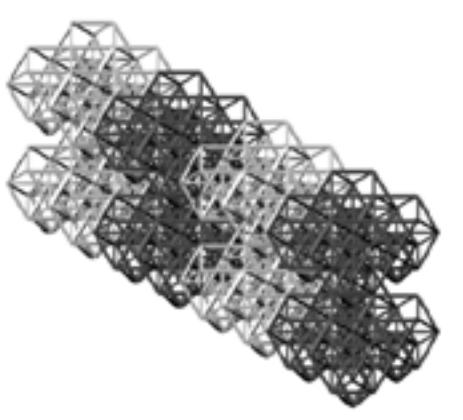
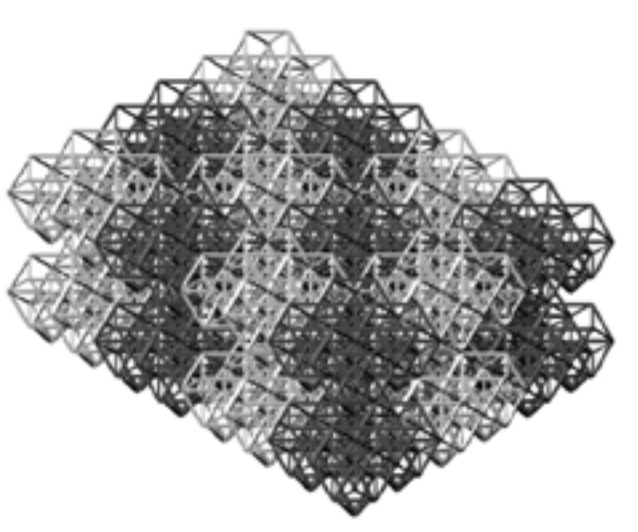
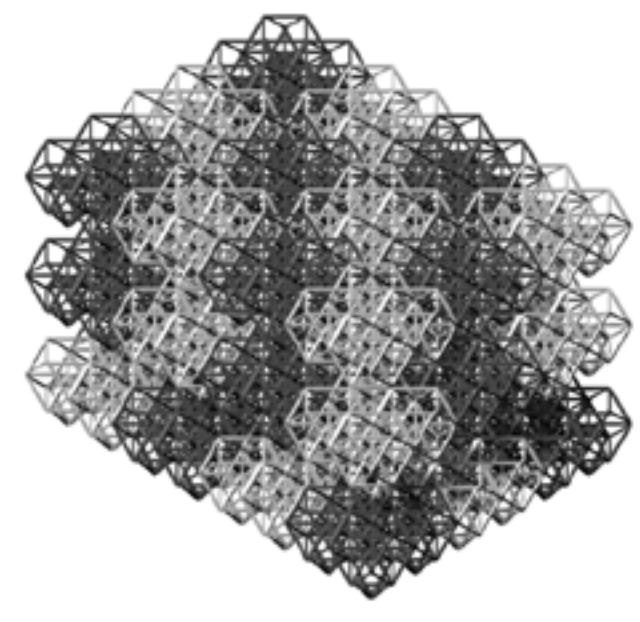

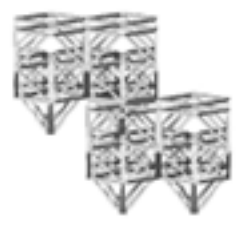
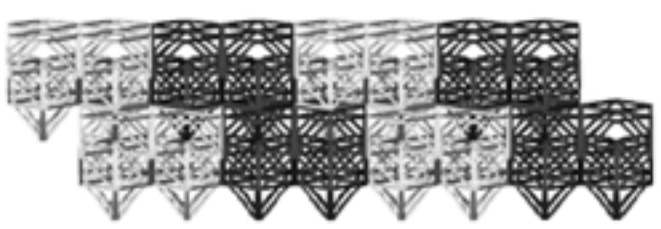
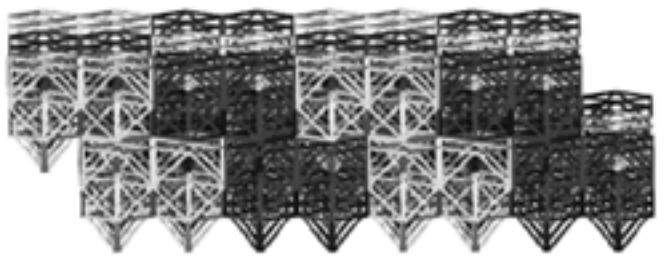
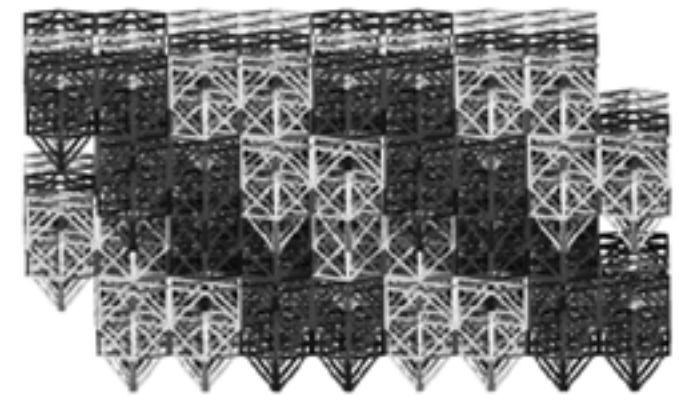

3

## INTRODUCTION

As robotic fabrication becomes increasingly integrated into architectural production, an emerging question is not only how machines can fabricate complex forms, but how computation can be embedded directly into material systems (Petersen et al. 2019; Leder and Menges 2024; Bagheri et al. 2025; Kyaw et al. 2024; Spencer et al. 2023). This project frames robotic assembly as a material practice in which geometry, structure, connection, and machine action are encoded into the assembly components (Fig.1), (Fig.2), (Fig.3).

2 Self aligning compound nested cuboctahedra–octahedral lattice unit with a reversible screw connector. These lattice-based units maintain geometric rigidity while remaining lightweight and reusable. The average density of the lattice is 81.85 grams per 100 mm³. The flanges in the modules enable support-free 3D printing.

3 The aggregation of the geometry allows the structure to grow along the x, y, and z axes, following the offset orientation of the compound nested module. Adjacent modules nest between the aggregated modules in the layer below, forming a staggered 3D interlocking assembly.

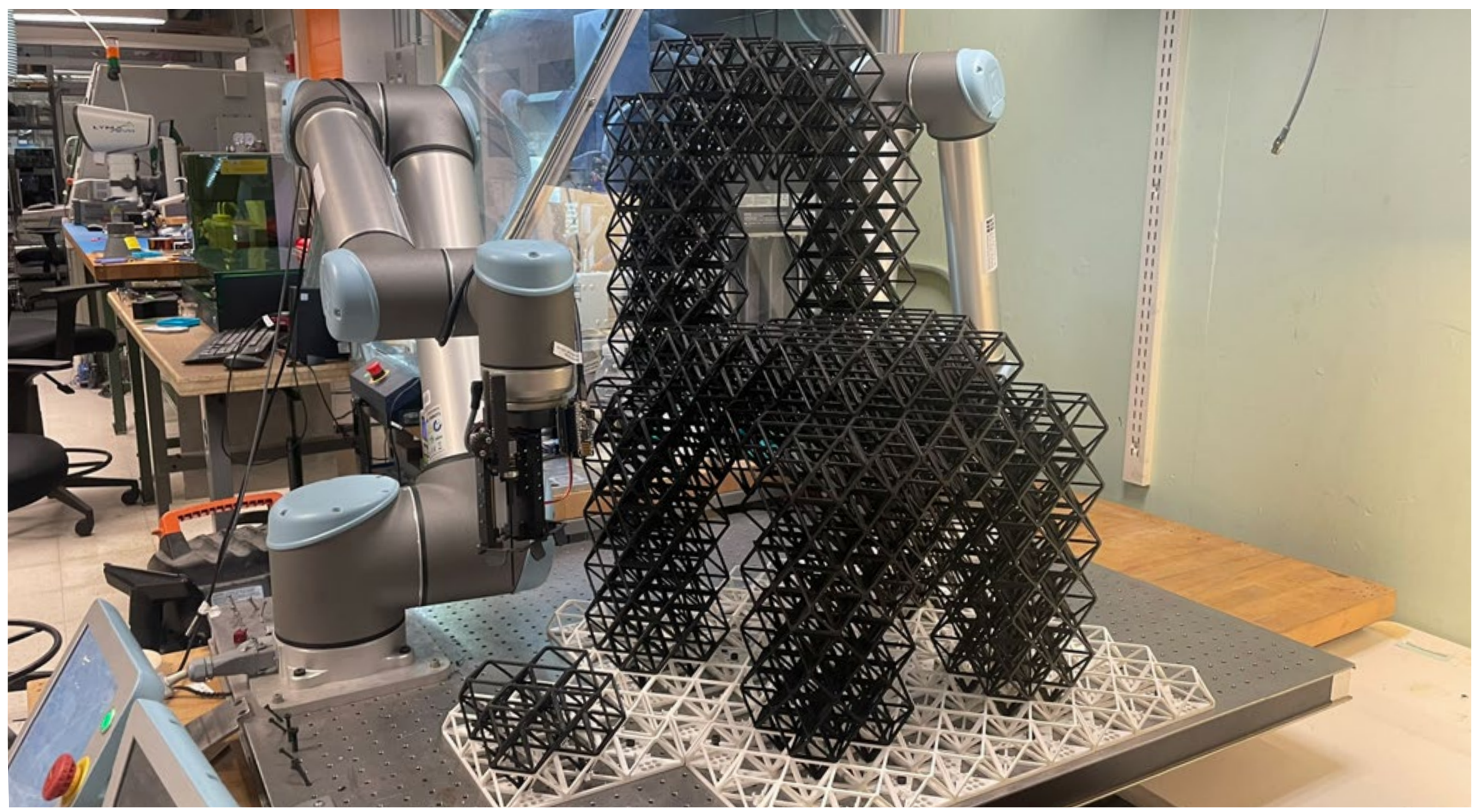

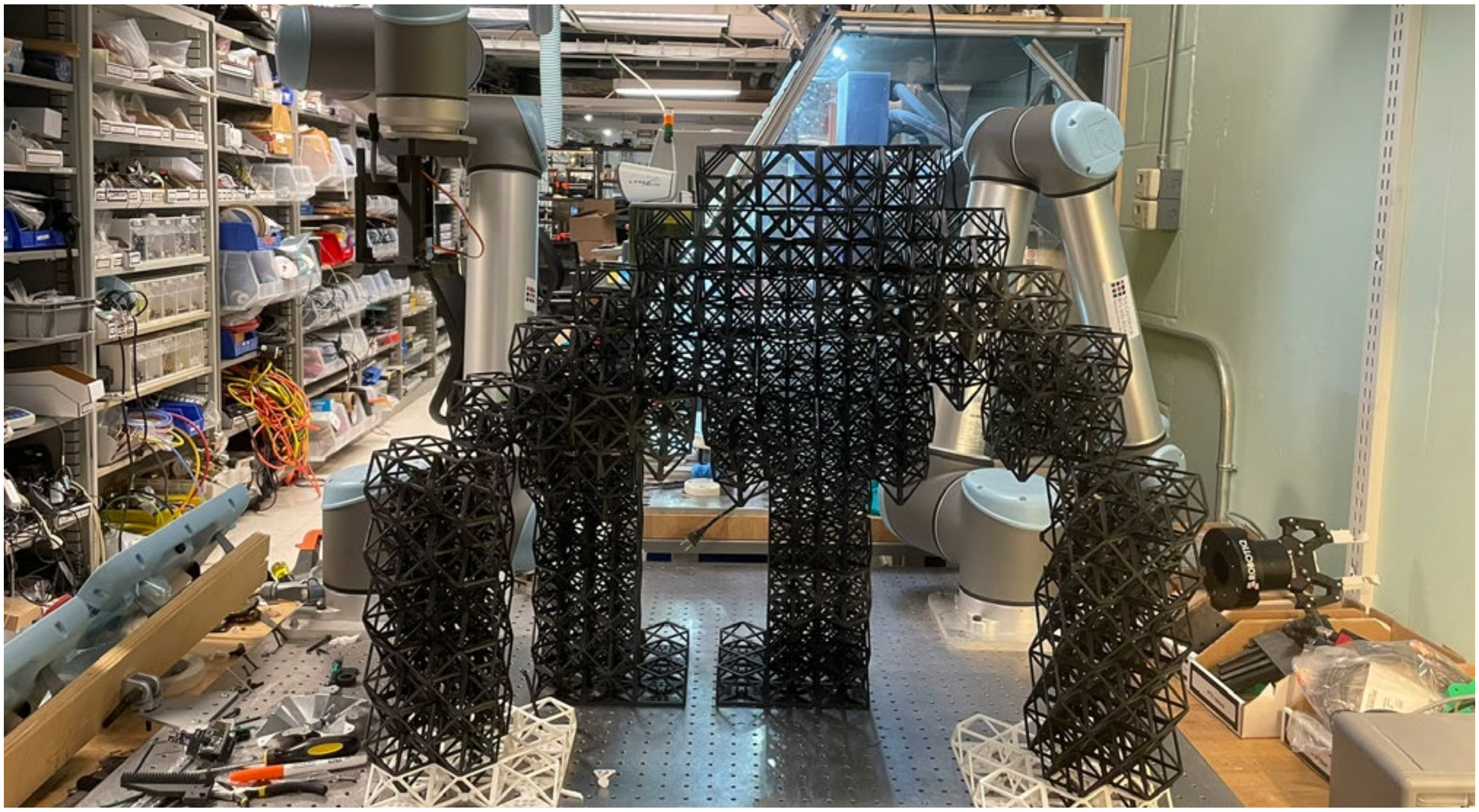

4

## BACKGROUND

Modular and discrete assembly systems are composed of units that can be repeatedly assembled, disassembled, and reconfigured into different built forms, making them especially relevant to circular construction, temporary structures, and adaptable architectural systems (Garusinghe et al. 2023; Kyaw, Jeon, et al. 2025; Mutis 2026; Smith et al. 2026).

To manage the complexity of assembling discrete components, prior work in robotic assembly has often focused on hardware design, motion planning, sensing, and vision systems (Felbrich et al. 2017; Hosmer et al. 2024; Mehrotra et al. 2026; Adel et al. 2026; Smith et al. 2025; Kyaw, Gupta, et al. 2025; Mutis et al. 2026). This project asks how material systems can embed structural logic, robotic grasping, self-alignment,

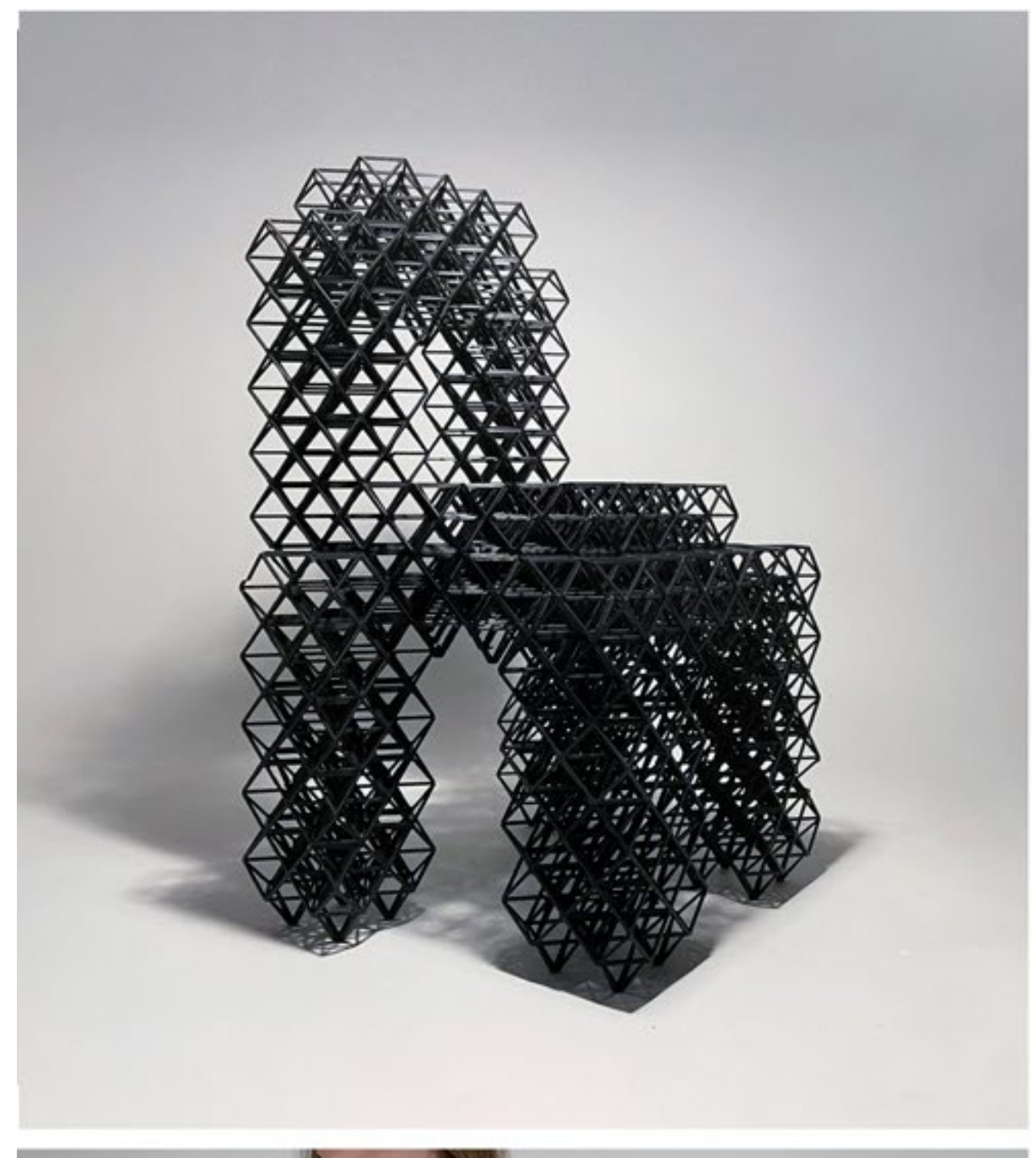
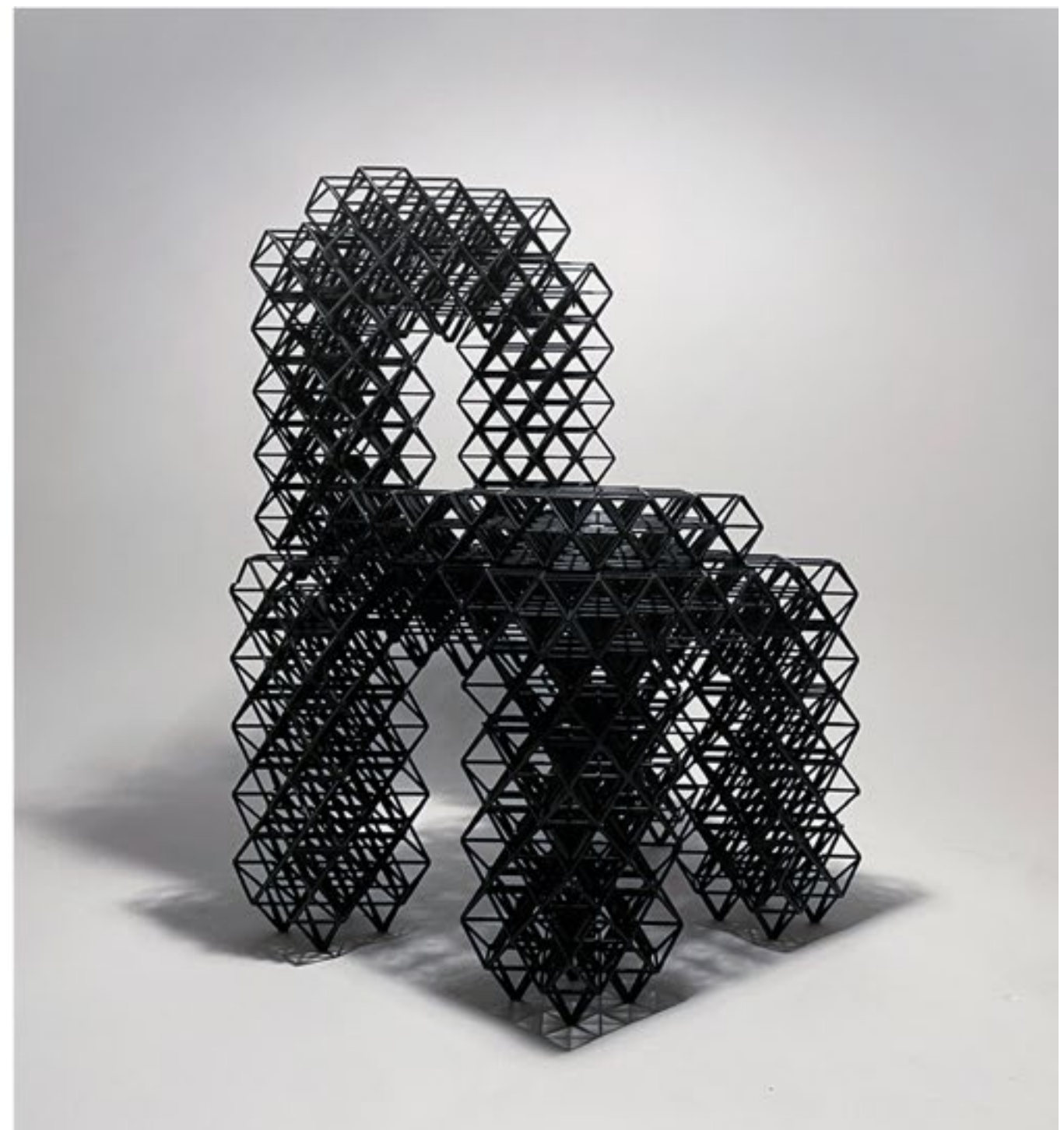
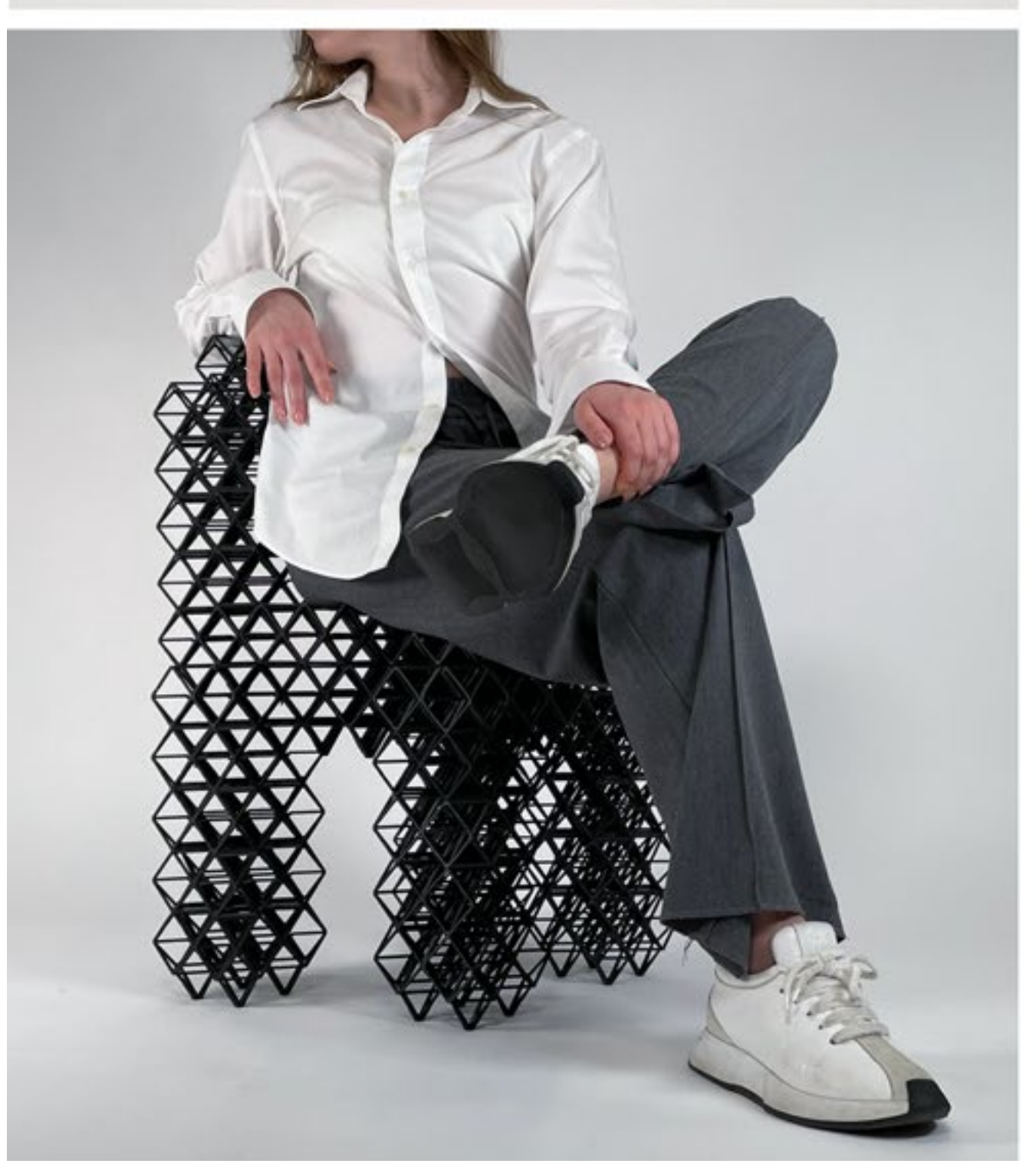
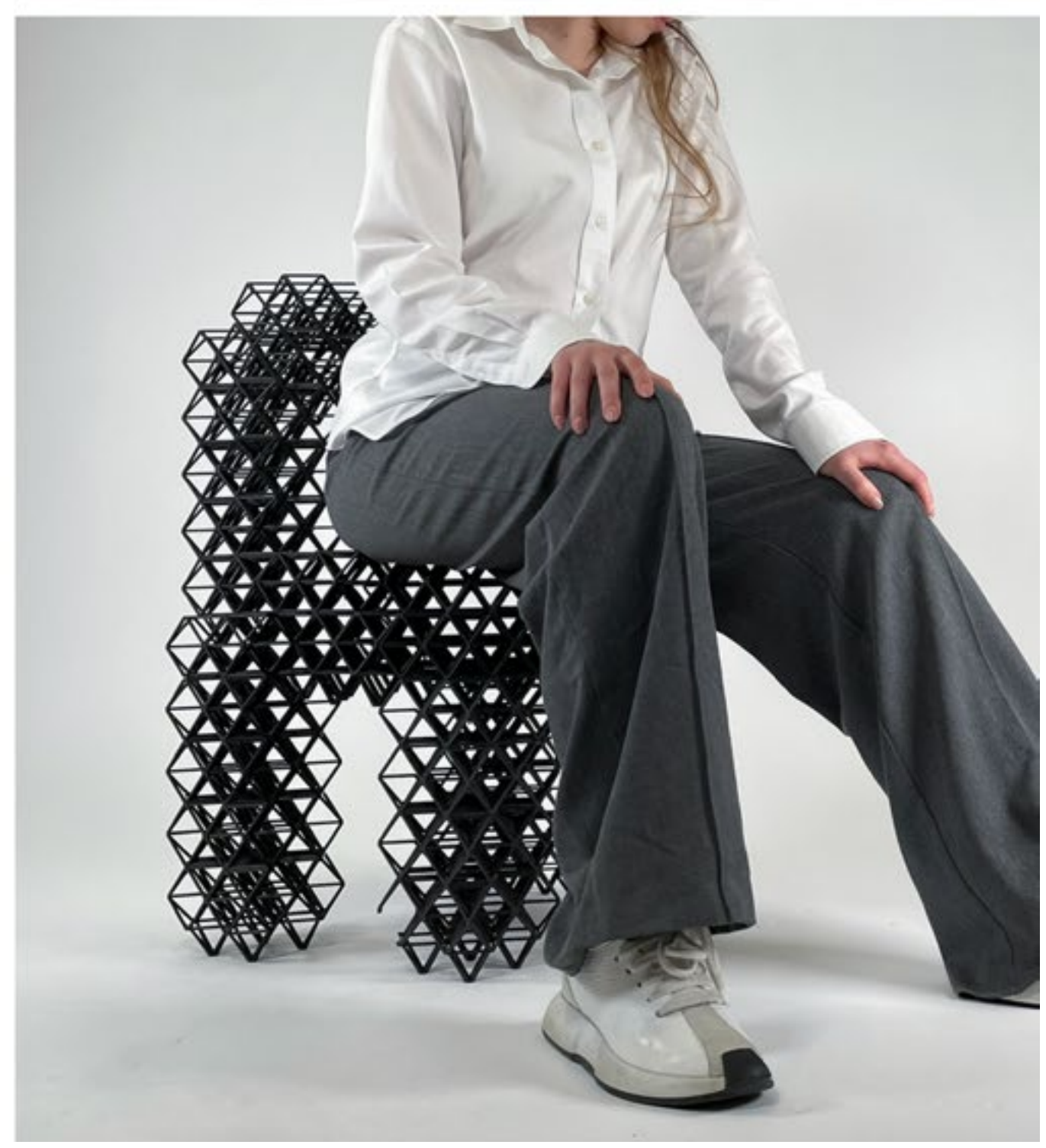

5

reversible connection, and material reuse directly into the form of an architected assembly module. More specifically, the goal is to co-design structural lattice modules and mobile robotic assemblers to construct scalable, reconfigurable, load-bearing structures (Fig 4). In this context, the module becomes more than a building block. It becomes a unit of computation, structure, and material intelligence.

4 Aggregation and assembly of a structural chair and a mound structure with multiple points of cantilevering, using a 6-axis industrial robotic arm and a custom gripper.

5 Assembled chair with a backrest supporting the full weight of the human body and leaning. The chair has a parallax visual effect as the voxel lattice's perspective changes.

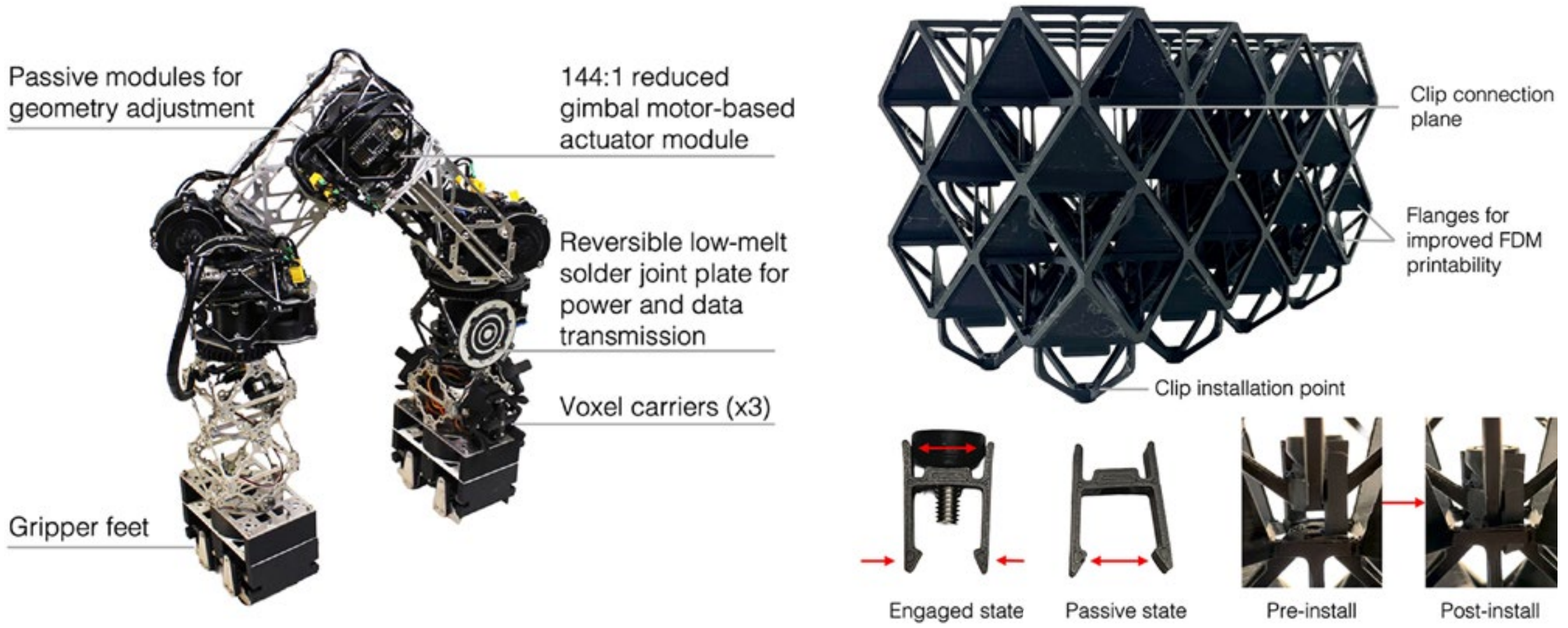

6

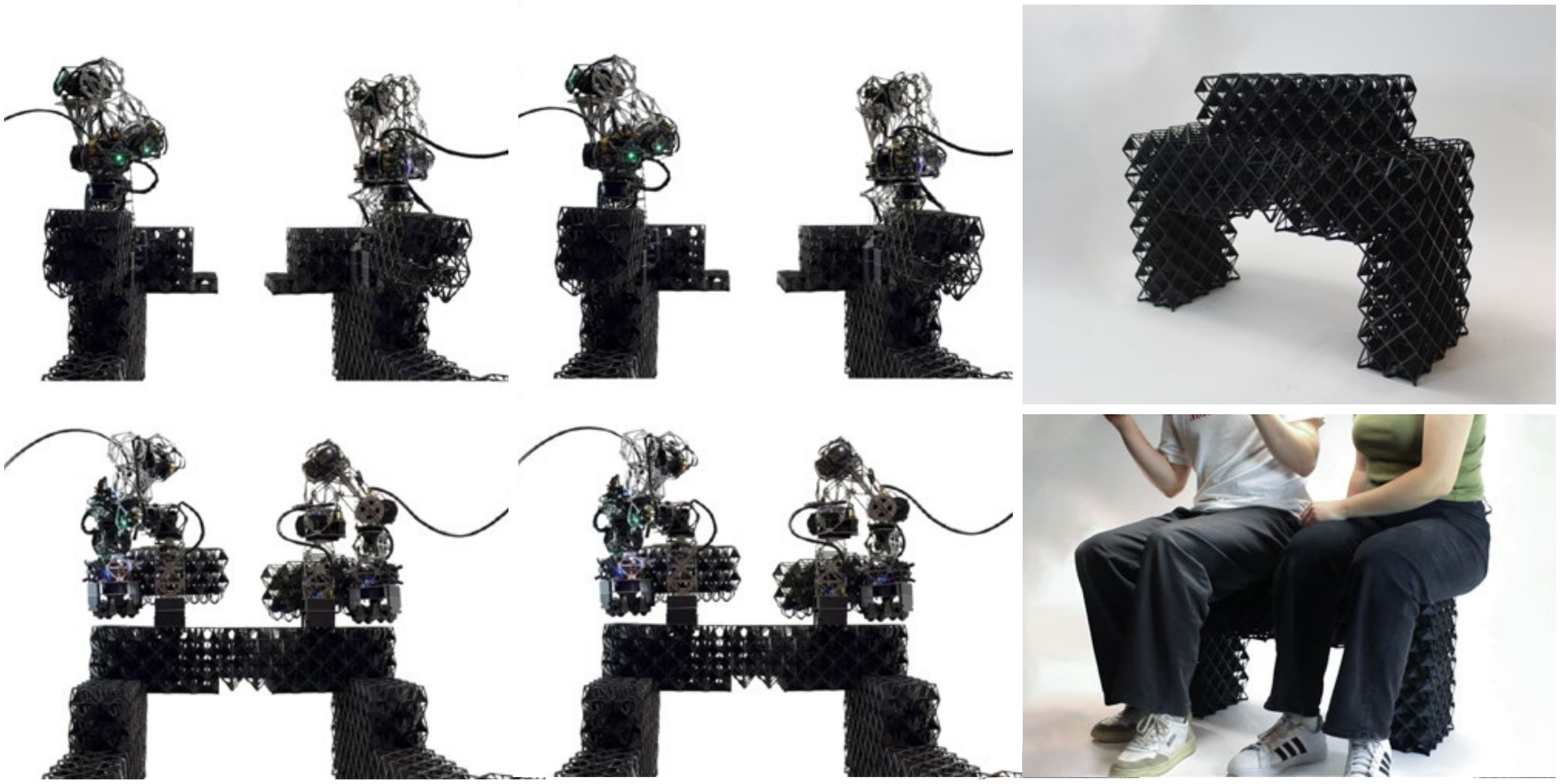

7

## METHODS

The system presents a self-aligning conjoined cuboctahedra-octahedral lattice unit (Fig.2). The cuboctahedra geometry provides a defined area for robotic grasping and functions as an alignment feature for the octahedral body of another unit to settle into. The octahedral feature is composed of two square pyramids joined at their bases. The lower square pyramid contains a screw reversible snap-fit connector, while the upper square pyramid contains the receptor.

This project introduces a novel compound nested module composed of 8 conjoined cuboctahedra-octahedral units to enable new form of interlocking assembly. 4 compounded units form the bottom layer, and 4 compounded units form an offset nested top layer (Fig.3). Within this compound nested module, 2 units connect the upper and lower compounded modules. The remaining 3 top units enable outward aggregation in the x, y, and z directions, while the 3 bottom units receive incoming connections from the x, y, and z directions. This staggered, offset, compound nested geometry creates an interlocking octet-based

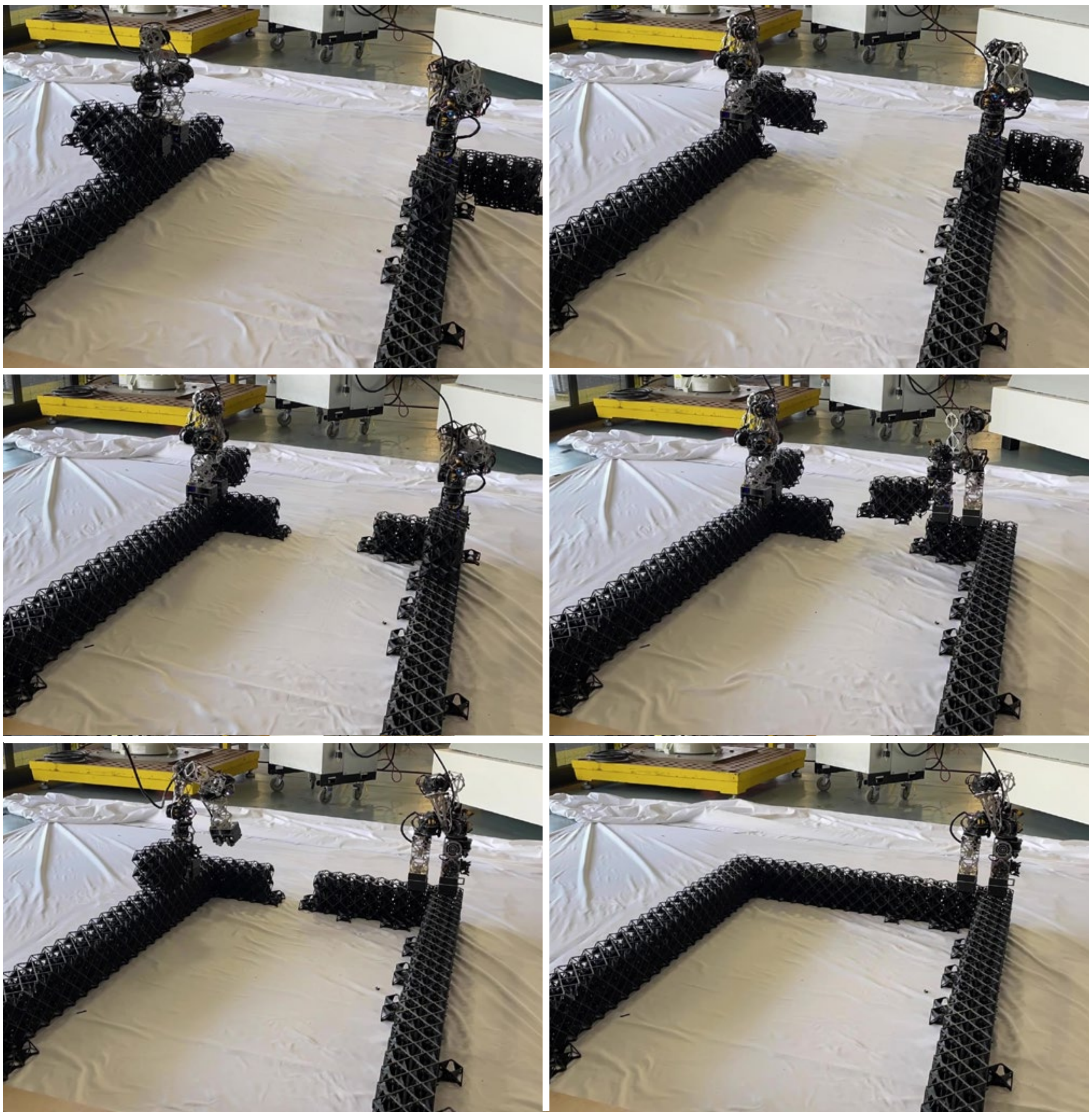

8

lattice assembly system that can aggregate in 3D with connections nested into adjacent modules (Fig.5). Therefore, the module informs the placement, alignment, connection, and aggregation of the assemblies

To demonstrate the system's modularity, aggregation logic, self-aligning feature, and feasibility, the project reveals several new built structures assembled through a robotic arm and a mobile assembler (Fig.6) (Fig.7) (Fig 8), These demonstrates that the modular system can support multiple architectural conditions, including seating (fig.5), spanning (fig.7), surface formation (fig.9), and vertical framing (fig.10).

6 Additional details of the mobile robotic assemblies and the reversible snap-fit connections of the compound nested lattice modules.

7 Step-by-step assembly of a bench using mobile robots, featuring two-person seating with a small backrest.

8 Two mobile robots assembling a door on the floor; the figure shows multiple steps of the assembly process.

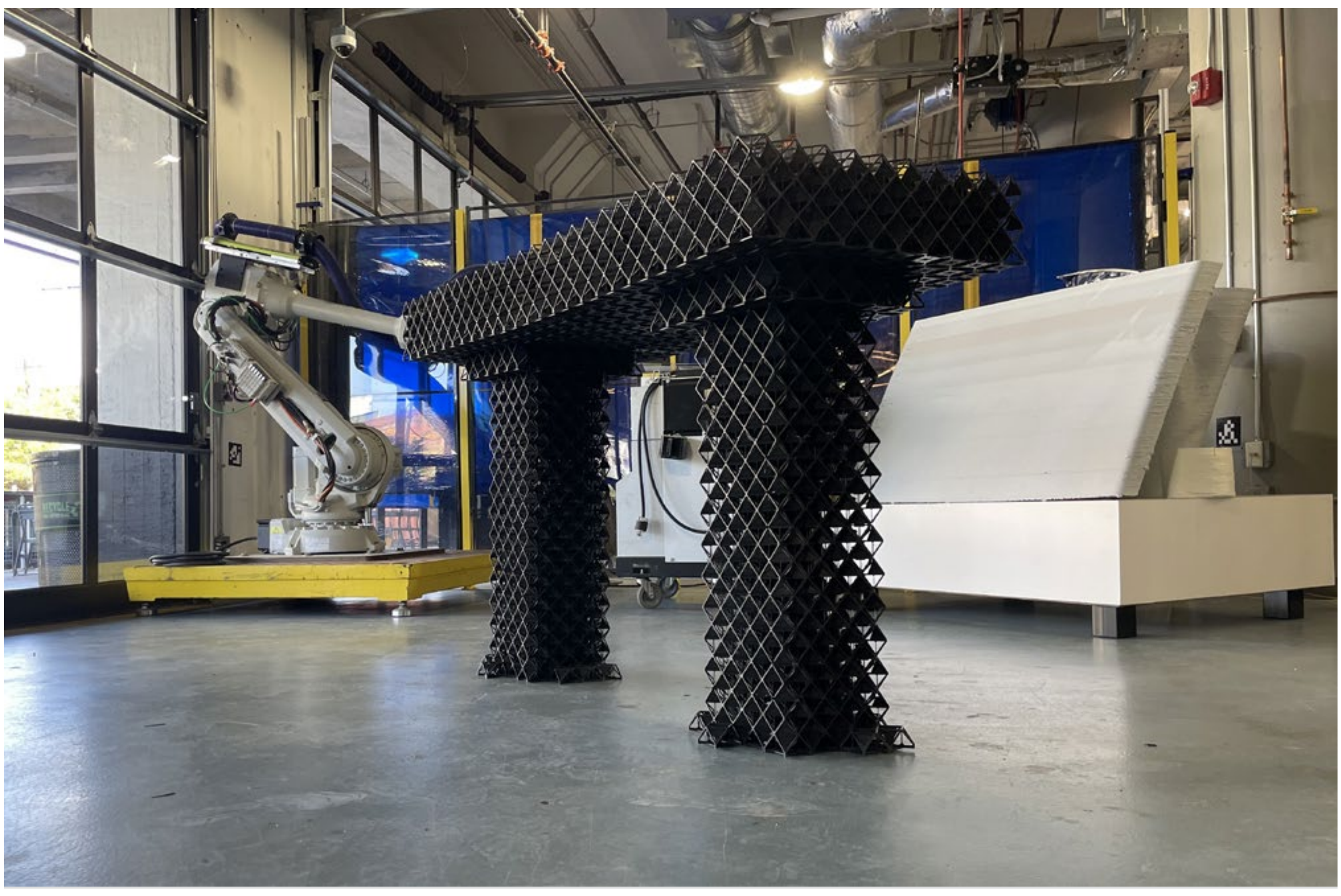

9

## RESULTS AND DEMONSTRATIONS

The resulting assemblies demonstrate that compound nested lattice modules in both furniture and architectural scales. The compounded module was tested in compression on an Instron 4411 equipped with a 5 kN load cell, resulting in a stiffness of 4,556 N/mm, a maximum load of 3,445 N, and a compressive modulus of 17.5 MPa.

Under nondestructive loading, the assembled chair supported over 150 pounds, the bench supported over 300 pounds, and the table supported over 450 pounds. The door supported its self weight during robotic assembly and during human assisted lifting to an upright position, maintaining its integrity in both tension and compression. It is later disassembled and reassembled into a table, demonstrating the capacity for circularity and component reusability.

## CONCLUSION AND FUTURE WORK

The geometry of the compound nested lattice module is encoded with instructions for how it should be gripped, aligned, connected, and aggregated. The project demonstrates that computation can be embedded beyond software, toolpaths, or robotic control. The module becomes an architected unit of computation that can be encoded with information. In doing so, the work recodes the relationship between geometric intelligence, structure, circularity, and robotic assembly. Future work can also explore the integration of mixed reality and natural language interfaces to enable humans and robots to better communicate in the process. (Kyaw 2023; Kyaw, Smith, et al. 2025)

## ACKNOWLEDGMENTS

This research is supported by CBA Consortia funding and the MIT Morningside Academy of Design.

9 Assembled table with two columns and a horizontally spanning surface aggregation, forming a self-supporting structure capable of supporting approximately 450 pounds.

10 A door frame that is robotically assembled flat on the ground and lifted into an upright position while supporting its own self-weight.

10

## IMAGE CREDITS

All drawings and images by the authors.

**Alexander Htet Kyaw** is a PhD student at MIT working at the intersection of artificial intelligence, augmented reality, robotics, design, and fabrication. He is currently affiliated with MIT CSAIL, the Media Lab, and the Morningside Academy of Design, and has previously worked at Google, Microsoft, Autodesk, SOM, and Jenny Sabin Studio. He holds master's degrees from MIT in Design Computation and Electrical Engineering and Computer Science, and completed his Bachelor of Architecture at Cornell University.

**Miana Smith** is a PhD student at MIT's Center for Bits and Atoms. Her research focuses on reconfigurable robotic systems for assembling large-scale structures from reusable discrete components. She develops scalable approaches to robotic construction that connect digital fabrication, modular design, and sustainable manufacturing. She holds a bachelor's degree in mechanical engineering and a master's degree in media arts and sciences from MIT.

**Paul Richard** is a master's student in robotics at EPFL and a former visiting researcher at MIT. His work focuses on robotic systems, motion control, and automated assembly, including methods for constructing large structures from reusable modular building blocks. He is interested in how robotics can support scalable and resource-efficient fabrication.

**Neil Gershenfeld** is a professor at MIT and director of the MIT Center for Bits and Atoms. His research spans digital fabrication, computing, and the relationship between information and physical form. He founded the global Fab Lab network and leads the Fab Academy, expanding access to digital fabrication around the world.